\documentclass[10pt,twocolumn,letterpaper]{article}

\usepackage{iccv}              

\definecolor{iccvblue}{rgb}{0.21,0.49,0.74}
\usepackage[pagebackref,breaklinks,colorlinks,allcolors=iccvblue]{hyperref}

\usepackage{hyperref}
\hypersetup{
colorlinks = true, 
urlcolor = magenta, 
linkcolor = red, 
citecolor = blue}
\usepackage{multirow}
\usepackage{enumitem}
\usepackage{romannum}
\usepackage{booktabs}
\usepackage{amsmath}
\usepackage{cuted}
\usepackage{capt-of}

\usepackage{array}
\newcolumntype{P}[1]{>{\centering\arraybackslash}p{#1}}

\usepackage{xcolor,colortbl}
\definecolor{Gray}{gray}{0.9}

\def\paperID{2} 
\def\confName{ICCV}
\def\confYear{2025}

\title{TransFlow: Motion Knowledge Transfer from Video Diffusion Models\\to Video Salient Object Detection}

\author{Suhwan Cho$^1$\quad Minhyeok Lee$^2$\quad Jungho Lee$^2$\quad Sunghun Yang$^2$\quad Sangyoun Lee$^2$\vspace{0.2cm}\\
$^1$~GenGenAI\quad $^2$~Yonsei University\vspace{0.1cm}\\
\fontsize{10.0}{10.0}\url{https://github.com/suhwan-cho/TransFlow}\\
}

\begin{document}
\maketitle

\begin{abstract}
Video salient object detection (SOD) relies on motion cues to distinguish salient objects from backgrounds, but training such models is limited by scarce video datasets compared to abundant image datasets. Existing approaches that use spatial transformations to create video sequences from static images fail for motion-guided tasks, as these transformations produce unrealistic optical flows that lack semantic understanding of motion. We present TransFlow, which transfers motion knowledge from pre-trained video diffusion models to generate realistic training data for video SOD. Video diffusion models have learned rich semantic motion priors from large-scale video data, understanding how different objects naturally move in real scenes. TransFlow leverages this knowledge to generate semantically-aware optical flows from static images, where objects exhibit natural motion patterns while preserving spatial boundaries and temporal coherence. Our method achieves improved performance across multiple benchmarks, demonstrating effective motion knowledge transfer.

\end{abstract}

\begin{figure}[t]
\centering
\includegraphics[width=0.84\linewidth]{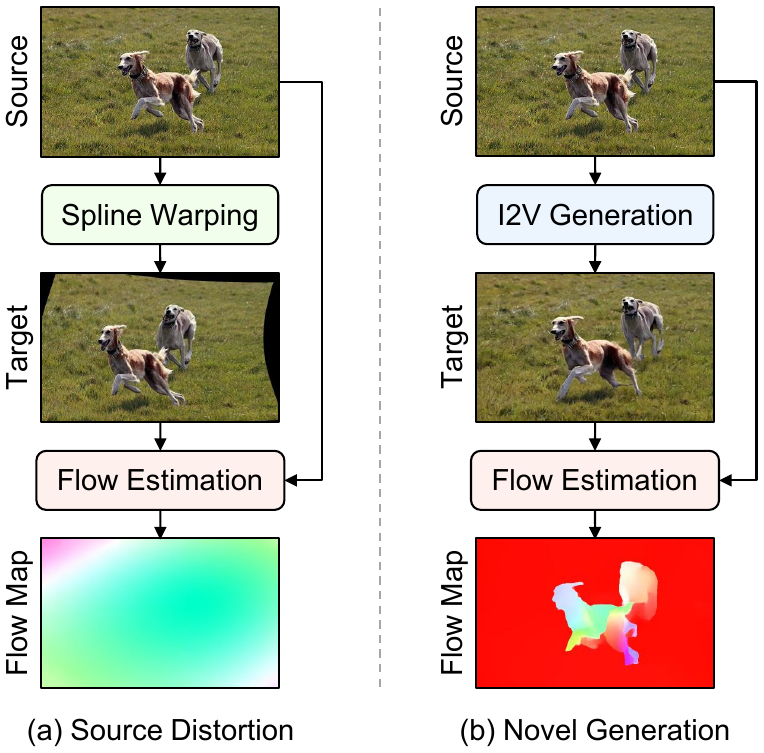}
\vspace{-1mm}
\caption{Motion generation comparison: (a) geometric transformations; (b) TransFlow's semantic motion generation.}
\label{figure1}
\end{figure}

\section{Introduction}
The success of deep learning in video processing relies heavily on large-scale datasets. While image datasets are abundant and relatively easy to annotate, collecting and labeling video data remains costly and labor-intensive. To address this challenge, researchers often simulate video sequences from image datasets using spatial transformations such as affine and spline warping. However, these methods fail for motion-guided tasks like video salient object detection (SOD), where realistic optical flow is essential for distinguishing salient objects from backgrounds.

Flow-guided video SOD models require optical flow maps to capture motion dynamics, but existing image datasets lack this crucial temporal information. While recent video segmentation datasets~\cite{mose, MeViS, survey} that have advanced the video understanding field can provide optical flow, their segmentation masks are misaligned with salient object annotations, as they focus on general object segmentation rather than salient object segmentation. This fundamental mismatch between available annotations and SOD requirements leaves the data scarcity challenge for video SOD largely unaddressed, consequently constraining model training and hindering both performance and generalization capabilities across diverse video scenarios.

The fundamental limitation of current flow simulation lies in its geometric approach to motion generation. Spatial transformations like affine and spline warping enforce uniform motion patterns across entire images, treating them as rigid entities rather than understanding that different objects should move independently according to their semantic properties. This produces unrealistic optical flows that lack meaningful motion cues, as illustrated in Figure~\ref{figure1}~(a).

To address this limitation, we propose TransFlow, a framework that transfers motion knowledge from pre-trained video diffusion models to generate realistic training data for video SOD. Unlike geometric transformations, video diffusion models have learned semantic motion understanding from large-scale video data, capturing how different objects naturally move in real scenes. TransFlow leverages this knowledge to generate semantically-aware optical flows from static images, where objects exhibit contextually-appropriate motion patterns while preserving spatial boundaries, as shown in Figure~\ref{figure1}~(b). This represents a paradigm shift from geometric to semantic motion generation, enabling the creation of training datasets with meaningful motion cues.

We create a large-scale dataset of image-flow-mask triplets derived from existing image SOD datasets using TransFlow. This allows flow-guided video SOD models to benefit from abundant image data while obtaining realistic motion supervision. Experimental results show improved performance on multiple benchmark datasets, suggesting that transferring motion knowledge from generative models can effectively address data limitations in motion-guided video understanding tasks.

Our main contributions are as follows:
\begin{itemize}[leftmargin=0.2in]
\item We propose TransFlow, a framework for transferring semantic motion knowledge from video diffusion models to enhance video SOD training data.

\item We demonstrate that video diffusion models can generate semantically-aware motion patterns from static images, producing realistic optical flows that capture independent object motion.

\item We explore cross-paradigm knowledge transfer from generative to discriminative video understanding, showing how motion knowledge can be effectively harvested from foundation models.
\end{itemize}

\begin{figure*}[t]
\centering
\includegraphics[width=0.9\linewidth]{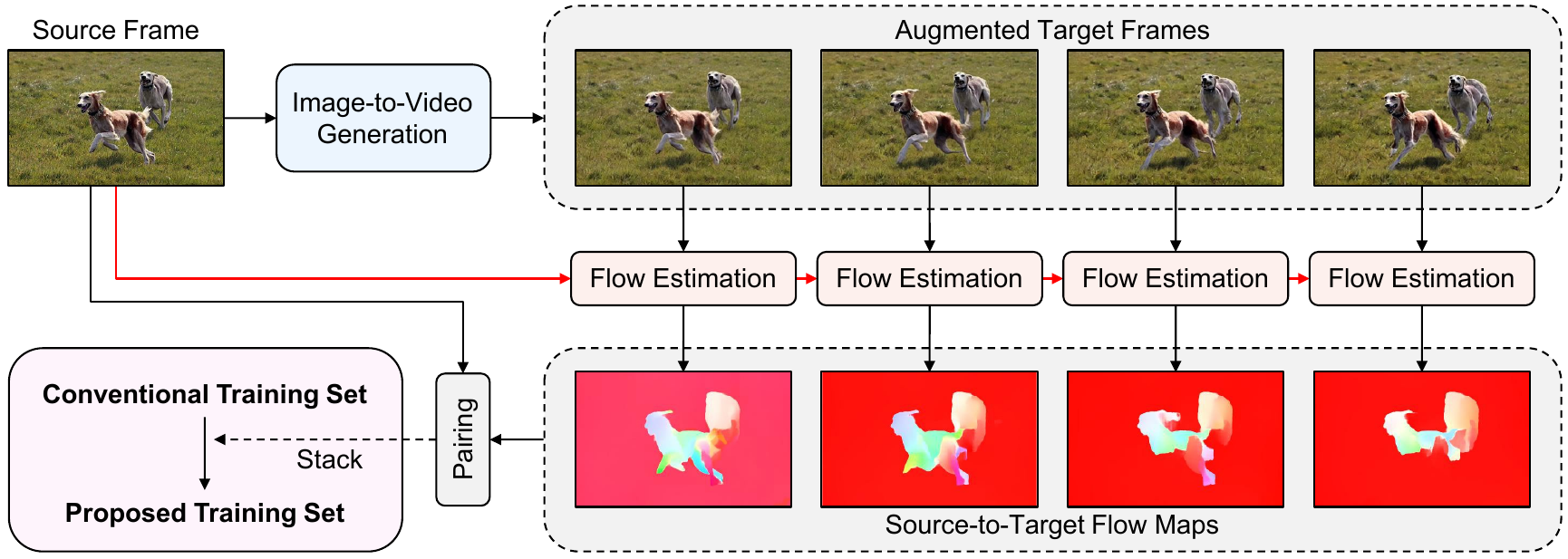}
\vspace{-1mm}
\caption{Training data generation pipeline. Static images are transformed into video sequences using diffusion models, then optical flows are estimated between source and generated frames to create image-flow training pairs.}
\label{figure2}
\end{figure*}

\section{Related Work}
\noindent\textbf{Utilizing image data for video models.}
To leverage the extensive knowledge embedded in large-scale image datasets, video processing tasks commonly incorporate image data through spatial transformations, treating original and transformed images as video sequences.

In video object segmentation (VOS), MaskTrack~\cite{MaskTrack} applies affine transformations and thin-plate spline~\cite{TPS} deformations to segmentation masks, using transformed masks as previous frame annotations alongside query frame images for training. RGMP~\cite{RGMP} extends this approach by overlaying foreground objects from salient object detection datasets~\cite{SOD1, SOD2} onto PASCAL-VOC~\cite{PASCAL1, PASCAL2} backgrounds while applying similar transformations. Recent methods~\cite{STM, KMN, HMMN, STCN, TBD, XMem, 3DC-Seg} generate video-like samples by independently transforming images and masks, then sequencing them temporally, demonstrating effectiveness across both semi- and unsupervised VOS settings.

Beyond VOS, STEm-Seg~\cite{STEm-Seg} synthesizes training clips for video instance segmentation using random affine transformations and motion blur effects, while FCNS~\cite{FCNS} employs similar augmentation techniques to enhance saliency learning and reduce overfitting.

However, these geometric transformation approaches treat images as rigid entities, applying uniform motion patterns without considering semantic properties of different objects. Consequently, the resulting optical flows lack meaningful motion cues that reflect how objects naturally move in real scenes, limiting their effectiveness for motion-guided tasks.

\vspace{1mm}
\noindent\textbf{Motion-guided architectures.}
Motion cues provide crucial insights into object dynamics, typically captured through optical flow estimation between consecutive frames. This temporal information is particularly valuable for segmentation tasks, where flow maps preserve spatial details that improve pixel-level prediction accuracy.

Recent methods integrate motion and appearance information through various fusion mechanisms. MATNet~\cite{MATNet} transforms appearance features into motion-attentive representations using asymmetric attention blocks, while MGA~\cite{MGA} employs separate processing modules for each modality and adapts them through end-to-end training. AMC-Net~\cite{AMC-Net} dynamically balances appearance and motion cues via gating functions, and FSNet~\cite{FSNet} enforces mutual constraints between multi-modal features during encoding. Methods such as PMN~\cite{PMN}, GSA-Net~\cite{GSA-Net}, and DPA~\cite{DPA} further leverage attention mechanisms to fuse cross-modal information while capturing temporal dependencies across video sequences.

Despite these architectural advances, flow-guided methods remain constrained by limited training data with realistic motion patterns. Traditional image-to-video simulation produces optical flows from geometric transformations that lack authentic motion characteristics, reducing their effectiveness for training robust motion-guided models.

\vspace{1mm}
\noindent\textbf{Image-to-video generation.}
Video generation models have gained significant attention for content creation and editing applications. Image-to-video generation models, which produce video sequences from single images, demonstrate superior quality and controllability by using input images as explicit first-frame guides.

Diffusion models~\cite{DDPM, DDIM, LDM} have significantly advanced image-to-video generation capabilities. Representative approaches include Stable Video Diffusion~\cite{SVD}, Gen-2~\cite{Gen-2}, I2VGen-XL~\cite{I2VGen-XL}, PikaLabs~\cite{PikaLabs}, SparseCtrl~\cite{SparseCtrl}, and SORA~\cite{SORA}. Some methods extend controllability through explicit optical flow or trajectory guidance~\cite{DragNUWA, MoVideo}.

Video diffusion models have learned rich semantic motion priors from large-scale video data, encoding understanding of how different objects naturally move in diverse scenes. However, existing work primarily focuses on content generation rather than leveraging this embedded motion knowledge for discriminative video understanding tasks. Our work explores transferring this motion understanding to generate realistic training data for video SOD.

\section{Approach}

\subsection{Generating Target Frames}
As shown in Figure~\ref{figure1}, spatially distorting the original source image to generate a target frame for optical flow estimation does not yield realistic optical flow maps. To obtain high-quality optical flow maps with video-like properties from static images, we use an image-to-video generation model that generates new frames directly, rather than relying on distortion of the source image.

\vspace{1mm}
\noindent\textbf{Network overview.}
In the image-to-video generation process, we employ Stable Video Diffusion~\cite{SVD} based on a 3D-UNet~\cite{3D-UNet} architecture. Following recent approaches, pixel-to-latent conversion is applied before the denoising model, with latent-to-pixel conversion following it, using a pre-trained VAE~\cite{VAE} encoder and decoder, respectively. The VAE decoder incorporates 3D convolutional layers to account for the temporal dimension, ensuring temporal consistency during the decoding process.
Starting from Gaussian noise $x \in \mathbb{R}^{C \times T \times H \times W}$ and the source image latent $z_s \in \mathbb{R}^{C \times H \times W}$, the denoised latent $z \in \mathbb{R}^{C \times T \times H \times W}$ is obtained as:
\begin{align}
z = \Phi(x, z_s^{(T)}),
\end{align}
where $\Phi$ denotes the denoising 3D-UNet model and $z_s^{(T)}$ represents the temporal replication of $z_s$ across $T$ frames. The output denoised latent $z$ is then decoded with the VAE decoder on a per-frame basis, with each frame representing a transformed image generated from the source image. The number of video frames $T$ is set to $14$, following the default setting in Stable Video Diffusion.

\vspace{1mm}
\noindent\textbf{Sampling details.}
During the sampling process, we use $25$ steps of the deterministic DDIM sampler~\cite{detDDIM}. The classifier guidance scale~\cite{CFG} is set to $3.0$ for the first frame and $1.0$ for the last frame. The resolution and frame rate are set to 576$\times$1024 and 7 fps, respectively, following the default settings. The decoding chunk size is set to $8$ to balance generation quality and computational cost.
\subsection{Hallucinating Flows}
After generating target images from each source image using the image-to-video generation model, we obtain sets of source-target pairs for triplet construction. The temporary paired data $P$ can be represented as:
\begin{align}
P_s = \left\{(I_s, I_t) : t \in \{1, 2, \ldots, T\}\right\},
\end{align}
where $I_s$ represents the source image and $I_t$ denotes the generated target image at frame $t$. From these temporary pairs, we construct complete training triplets for network training, as illustrated in Figure~\ref{figure2}.

\vspace{1mm}
\noindent\textbf{Triplet creation.}
To obtain the final training triplets for the two-stream video SOD network, we estimate optical flows from the temporary paired data $P$ and leverage the corresponding ground-truth masks. For optical flow estimation, we use a pre-trained RAFT model~\cite{RAFT} while maintaining the original image resolution. The final training triplets $Q$ can be obtained as:
\begin{align}
Q_s = \left\{(I_s, F_{s \to t}, M_s) : t \in \{1, 2, \ldots, T\}\right\},
\end{align}
where $F_{s \to t}$ indicates the estimated optical flow map from the source frame $s$ to target frame $t$, and $M_s$ represents the ground-truth saliency mask corresponding to the source image $I_s$. Since optical flows are computed using the source image as the reference frame, the RGB images, segmentation masks, and optical flows maintain perfect spatial alignment. The triplet data $Q$ is constructed for each static image separately, and therefore, from $N$ source images, we can obtain $NT$ complete training triplets.

\begin{table}[t!]
\centering 
\small
\caption{Training data comparison highlighting the scale advantages of diffusion-based data generation.}
\vspace{-2mm}
\begin{tabular}{c|c|c}
\toprule
Dataset &\#Triplets &\#Visual Contexts\\
\midrule
DAVIS 2016~\cite{DAVIS} &2,079 &30\\
DAVSOD~\cite{SSAV} &7,183 &61\\
DUTS-Video &218,008 &15,572\\
\bottomrule
\end{tabular}
\label{table1}
\end{table}

\vspace{1mm}
\noindent\textbf{Dataset construction.}
We apply our triplet creation protocol to the DUTS~\cite{DUTS} dataset, which consists of 10,553 training images and 5,019 testing images. Since these images are used solely for generating training data, we utilize the entire dataset (15,572 images total) for our pipeline. For each static image, we generate $T=14$ target frames using video diffusion models, then extract optical flows between the source and each target frame, resulting in 218,008 training triplets (15,572$\times$14). As shown in Table~\ref{table1}, this represents a substantial increase in scale compared to existing video SOD datasets: our DUTS-Video dataset contains 30$\times$ more triplets than DAVSOD~\cite{SSAV} (7,183 triplets) and 105$\times$ more than DAVIS 2016~\cite{DAVIS} (2,079 triplets), while providing 255$\times$ more visual contexts (15,572) than DAVSOD (61) and 519$\times$ more than DAVIS (30). These generated triplets provide complete image-flow-mask supervision and are combined with conventional video training data to enhance data diversity and provide realistic motion supervision for flow-guided video SOD training.

\subsection{Segmentation Model}
\begin{figure}[t]
\centering
\includegraphics[width=0.83\linewidth]{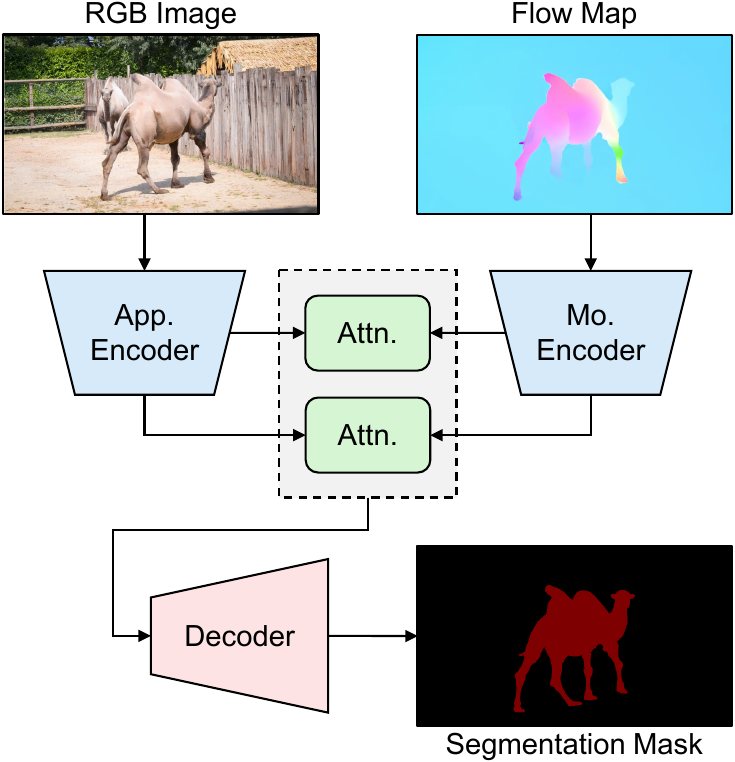}
\vspace{-1mm}
\caption{Two-stream network architecture for video SOD. The model processes RGB images and optical flow maps through separate encoders, fuses multi-level features via attention blocks, and outputs salient object masks.}
\label{figure3}
\end{figure}

We train a video SOD network using our generated image-flow-mask triplets in a fully supervised manner. Each training triplet consists of an RGB image, its corresponding optical flow map, and ground-truth saliency mask, enabling end-to-end supervised learning. Following conventional flow-guided approaches, we employ a two-stream architecture that processes appearance and motion information separately, as illustrated in Figure~\ref{figure3}.

\vspace{1mm}
\noindent\textbf{Architecture details.}
Our network employs dual MiT~\cite{MiT} encoders: one for RGB appearance features and another for optical flow motion features. Multi-level features from both streams are fused through CBAM attention blocks~\cite{CBAM}, then progressively decoded using feature interpolation with skip connections. The decoder outputs binary segmentation masks for salient object detection, supervised by ground-truth annotations.

\vspace{1mm}
\noindent\textbf{Training configuration.}
The network is trained in a fully supervised fashion using our triplet-based training data. We employ cross-entropy loss and Adam optimizer~\cite{adam} with learning rate 1e-5, batch size 16, and input resolution 512$\times$512. Training is performed on two GeForce RTX TITAN GPUs for approximately two days. We combine real video triplets with our generated training triplets: real data includes DAVIS 2016~\cite{DAVIS} and DAVSOD~\cite{SSAV} training sets, while synthetic data comes from our pipeline applied to DUTS images. Training triplets are randomly shuffled with a 2:1:1 mixture ratio for generated data, DAVIS 2016, and DAVSOD, respectively, ensuring balanced exposure to both diverse synthetic motion patterns and real video dynamics while maintaining complete supervised learning.

\begin{table*}[t]
\centering 
\caption{Quantitative evaluation on the DAVIS 2016 validation set, FBMS test set, DAVSOD test set, and ViSal dataset.}
\vspace{-2mm}
\small
\begin{tabular}{p{2.0cm}|P{5.8mm}P{5.8mm}P{5.8mm}|P{5.8mm}P{5.8mm}P{5.8mm}|P{5.8mm}P{5.8mm}P{5.8mm}|P{5.8mm}P{5.8mm}P{5.8mm}|P{5.8mm}P{5.8mm}P{5.8mm}}
\toprule
\multirow{2}{*}{Method} & \multicolumn{3}{c|}{DAVIS 2016} & \multicolumn{3}{c|}{FBMS} & \multicolumn{3}{c|}{DAVSOD} & \multicolumn{3}{c|}{ViSal} & \multicolumn{3}{c}{Average} \\
 & $\mathcal{S}\uparrow$ & $\mathcal{F}\uparrow$ & $\mathcal{M}\downarrow$ & $\mathcal{S}\uparrow$ & $\mathcal{F}\uparrow$ & $\mathcal{M}\downarrow$ & $\mathcal{S}\uparrow$ & $\mathcal{F}\uparrow$ & $\mathcal{M}\downarrow$ & $\mathcal{S}\uparrow$ & $\mathcal{F}\uparrow$ & $\mathcal{M}\downarrow$  & $\mathcal{S}\uparrow$ & $\mathcal{F}\uparrow$ & $\mathcal{M}\downarrow$ \\
\midrule
SSAV~\cite{SSAV} &89.3 &86.1 &2.8 &87.9 &86.5 &4.0 &72.4 &60.3 &9.2 &94.3 &93.9 &2.0 &86.0 &81.7 &4.5\\
F$^3$Net~\cite{F3Net} &85.0 &81.9 &4.1 &85.3 &81.9 &6.8 &68.9 &56.4  &11.7 &87.4 &90.7 &4.5 &81.7 &77.7 &6.8\\
MINet~\cite{MINet} &86.1 &83.5 &3.9 &84.9 &81.7 &6.7 &70.4 &58.2 &10.3 &90.3 &91.1 &4.1 &82.9 &78.6 &6.3\\
GateNet~\cite{GateNet} &86.9 &84.6 &3.6 &85.7 &83.2 &6.5 &70.1 &57.8 &10.4 &92.1 &92.8 &3.9 &83.7 &79.6 &6.1\\
PCSA~\cite{PCSA} &90.2 &88.0 &2.2 &86.6 &83.1 &4.1 &74.1 &65.5 &8.6 &94.6 &94.0 &1.7 &86.4 &82.7 &4.2\\
DFNet~\cite{DFNet} &- &89.9 &1.8 &- &83.3 &5.4 &- &- &- &- &92.7 &1.7 &- &- &-\\
3DC-Seg~\cite{3DC-Seg} &- &91.8 &1.5 &- &84.5 &4.8 &- &- &- &- &92.2 &1.9 &- &- &-\\
CASNet~\cite{CASNet} &87.3 &86.0 &3.2 &85.6 &86.3 &5.6 &69.4 &- &8.9 &82.0 &84.7 &2.9 &81.1 &- &5.2\\
FSNet~\cite{FSNet} &92.0 &90.7 &2.0 &89.0 &88.8 &4.1 &77.3 &68.5 &7.2 &- &- &- &- &- &-\\
CFAM~\cite{CFAM} &91.8 &90.9 &1.5 &90.9 &\underline{91.5} &\underline{2.6} &75.3 &66.2 &8.3 &94.7 &95.1 &1.3 &88.2 &85.9 &3.4\\
UFO~\cite{UFO} &91.8 &90.6 &1.5 &89.1 &88.8 &3.1 &- &- &- &\underline{95.9} &95.1 &1.3 &- &- &-\\
DBSNet~\cite{DBSNet} &92.4 &91.4 &1.4 &88.2 &88.5 &3.8 &77.8 &68.8 &7.6 &93.1 &92.8 &2.0 &87.9 &85.4 &3.7\\
HFAN~\cite{HFAN} &93.4 &92.9 &\textbf{0.9} &87.5 &84.9 &3.3 &75.3 &68.0 &7.0 &94.1 &93.5 &\underline{1.1} &87.6 &84.8 &3.1\\
TMO~\cite{TMO} &92.8 &92.0 &\textbf{0.9} &88.6 &88.2 &3.1 &76.7 &70.8 &7.2 &94.2 &94.7 &\textbf{1.0} &88.1 &86.4 &3.1\\
OAST~\cite{OAST} &93.5 &92.6 &1.1 &\underline{91.7} &\textbf{91.9} &\textbf{2.5} &78.6 &71.2 &7.0 &94.8 &95.0 &\textbf{1.0} &89.7 &87.7 &\underline{2.9}\\
TGFormer~\cite{TGFormer} &93.2 &92.2 &1.1 &91.6 &\textbf{91.9} &\underline{2.6} &\underline{79.8} &\underline{72.8} &\textbf{6.5} &95.2 &\underline{95.5} &\underline{1.1} &\underline{90.0} &\underline{88.1} &\textbf{2.8}\\
SimulFlow~\cite{simulflow} &\underline{93.7} &\underline{93.6} &\textbf{0.9} &- &- &- &77.1 &72.2 &6.9 &94.6 &94.3 &1.2 &- &- &-\\
\midrule
\textbf{TransFlow} &\textbf{94.5} &\textbf{93.9} &\underline{1.0} &\textbf{92.6} &90.6 &2.8 &\textbf{80.3} &\textbf{73.2} &\underline{6.6} &\textbf{96.2} &\textbf{96.6} &\textbf{1.0} &\textbf{90.9} &\textbf{88.6} &\underline{2.9}\\
\bottomrule
\end{tabular}
\label{table2}
\end{table*}

\section{Experiments}
We evaluate TransFlow on multiple video SOD benchmarks and conduct ablation studies to validate our approach. All inference experiments are performed on a single GeForce RTX 2080 Ti GPU.

\subsection{Datasets}
We utilize existing image and video SOD datasets for training and evaluation. For video datasets, optical flows are precomputed between consecutive frames using RAFT~\cite{RAFT} and stored alongside RGB images and segmentation masks.

\vspace{1mm}
\noindent\textbf{DAVIS 2016}~\cite{DAVIS}. This primary video SOD benchmark contains 50 videos with frame-level annotations, split into 30 training and 20 validation videos. We use the training set for model training and the validation set for evaluation.

\vspace{1mm}
\noindent\textbf{FBMS}~\cite{FBMS}. This dataset contains 59 videos (29 training, 30 testing) with sparse frame annotations. Following standard video SOD protocols, we use only the test set for evaluation.

\vspace{1mm}
\noindent\textbf{DAVSOD}~\cite{SSAV}. A recent video SOD benchmark comprising 61 training and 35 test videos with complete frame-level annotations. We use the training set for model training and the test set for evaluation.

\vspace{1mm}
\noindent\textbf{ViSal}~\cite{ViSal}. This dataset contains 17 videos with 193 annotated frames. Since it lacks a defined train/test split, we use it solely for evaluation.

\vspace{1mm}
\noindent\textbf{DUTS}~\cite{DUTS}. A large-scale image SOD dataset containing 10,553 training and 5,019 test images with pixel-level annotations. We use the entire dataset (15,572 images) for generating our DUTS-Video training data.

\vspace{1mm}
\noindent\textbf{DUTS-Video.} Our generated dataset created by applying our motion generation pipeline to DUTS images. From 15,572 static images, we generate 218,008 image-flow-mask triplets with semantically-aware motion patterns. This dataset is used exclusively for training TransFlow, enabling the model to benefit from realistic motion cues derived from abundant image data, effectively addressing the video data scarcity challenge.

\subsection{Evaluation Metrics}
We adopt three standard evaluation metrics for quantitative validation: $\mathcal{S}$-measure~\cite{smeasure}, $\mathcal{F}$-measure, and $\mathcal{M}$-measure.

\vspace{1mm}
\noindent\textbf{$\mathcal{S}$-measure.} The $\mathcal{S}$-measure evaluates structural similarity between predicted saliency maps and ground truth masks:
\begin{align}
&\mathcal{S} = \alpha \mathcal{S}_o + (1 - \alpha) \mathcal{S}_r~,
\end{align}
where $\mathcal{S}_o$ and $\mathcal{S}_r$ represent object-aware and region-aware structural similarity, respectively. We set $\alpha = 0.5$ following standard practice.

\vspace{1mm}
\noindent\textbf{$\mathcal{F}$-measure.} The $\mathcal{F}$-measure computes the weighted harmonic mean of precision and recall:
\begin{align}
&\mathcal{F} = \frac{(1 + \beta^2) \times \text{Precision} \times \text{Recall}}{\beta^2 \times \text{Precision} + \text{Recall}}~,
\end{align}
where $\beta^2 = 0.3$ following~\cite{fmeasure}.

\vspace{1mm}
\noindent\textbf{$\mathcal{M}$-measure.} The $\mathcal{M}$-measure computes the mean absolute error between predictions and ground truth:
\begin{align}
&\mathcal{M} = \frac{1}{H \times W} \sum_{p=1}^H \sum_{q=1}^W |M_{pred}^{p,q} - M_{gt}^{p,q}|~,
\end{align}
where $M_{pred}$ and $M_{gt}$ denote predicted and ground truth masks, respectively.

\begin{figure*}[t]
\centering
\includegraphics[width=1.0\linewidth]{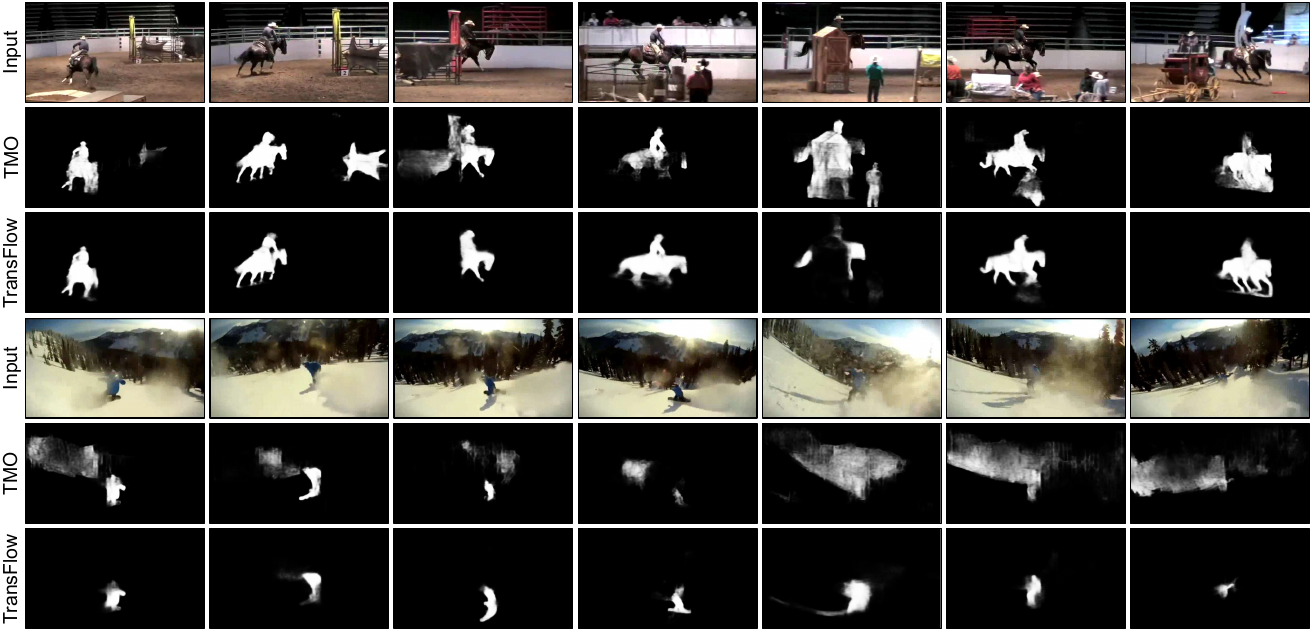}
\vspace{-6mm}
\caption{Qualitative comparison under challenging scenarios.}
\label{figure4}
\end{figure*}

\begin{table*}[t]
\centering 
\caption{Ablation study on the training protocol.}
\vspace{-2mm}
\small
\begin{tabular}{P{1.7cm}|P{5.8mm}P{5.8mm}P{5.8mm}|P{5.8mm}P{5.8mm}P{5.8mm}|P{5.8mm}P{5.8mm}P{5.8mm}|P{5.8mm}P{5.8mm}P{5.8mm}|P{5.8mm}P{5.8mm}P{5.8mm}}
\toprule
\multirow{2}{*}{Training} & \multicolumn{3}{c|}{DAVIS 2016} & \multicolumn{3}{c|}{FBMS} & \multicolumn{3}{c|}{DAVSOD} & \multicolumn{3}{c|}{ViSal} & \multicolumn{3}{c}{Average} \\
 & $\mathcal{S}\uparrow$ & $\mathcal{F}\uparrow$ & $\mathcal{M}\downarrow$ & $\mathcal{S}\uparrow$ & $\mathcal{F}\uparrow$ & $\mathcal{M}\downarrow$ & $\mathcal{S}\uparrow$ & $\mathcal{F}\uparrow$ & $\mathcal{M}\downarrow$ & $\mathcal{S}\uparrow$ & $\mathcal{F}\uparrow$ & $\mathcal{M}\downarrow$  & $\mathcal{S}\uparrow$ & $\mathcal{F}\uparrow$ & $\mathcal{M}\downarrow$ \\
\midrule
Real &93.1 &91.6 &1.3 &85.8 &85.5 &5.3 &76.0 &68.7 &7.8 &94.5 &94.4 &1.8 &87.4 &85.1 &4.1\\
Synthetic &91.7 &89.6 &2.0 &91.0 &89.0 &2.9 &78.1 &70.1 &7.8 &96.1 &96.4 &1.1 &89.2 &86.3 &3.5\\
Mixed &94.5 &93.9 &1.0 &92.6 &90.6 &2.8 &80.3 &73.2 &6.6 &96.2 &96.6 &1.0 &90.9 &88.6 &2.6\\
\bottomrule
\end{tabular}
\label{table3}
\end{table*}

\subsection{Quantitative Results}
We evaluate TransFlow against state-of-the-art methods on four standard benchmarks: DAVIS 2016~\cite{DAVIS} validation set, FBMS~\cite{FBMS} test set, DAVSOD~\cite{SSAV} test set, and ViSal~\cite{ViSal} dataset. As shown in Table~\ref{table2}, TransFlow achieves the highest $\mathcal{S}$-measure~\cite{smeasure} scores across all benchmarks, demonstrating superior structural similarity preservation. For $\mathcal{F}$-measure, our method ranks first on three datasets and remains competitive on the fourth, while maintaining comparable $\mathcal{M}$-measure performance.

The consistent improvements across diverse datasets validate that our semantic motion generation approach meaningfully enhances video SOD performance. These results demonstrate that addressing training data scarcity through large-scale synthesis is more effective than architectural innovations for advancing video SOD capabilities. Our findings suggest that data abundance, rather than model sophistication, represents the primary bottleneck in current video SOD development.

\subsection{Qualitative Results}
Figure~\ref{figure4} compares TransFlow with TMO~\cite{TMO} on challenging scenarios featuring complex background clutter and severe occlusion. While TMO produces fragmented segmentations and misses salient regions, TransFlow delivers complete and accurate object boundaries despite significant visual ambiguities and distractors.

This performance advantage stems from our large-scale synthetic training data, which exposes the model to diverse appearance-motion combinations unavailable in limited real datasets. The significant increase in training triplets compared to existing video SOD datasets enables robust feature learning and superior generalization to challenging real-world scenarios, demonstrating that synthetic data generation can effectively overcome the fundamental data scarcity bottleneck in video SOD.

\begin{figure*}[t]
\centering
\includegraphics[width=1.0\linewidth]{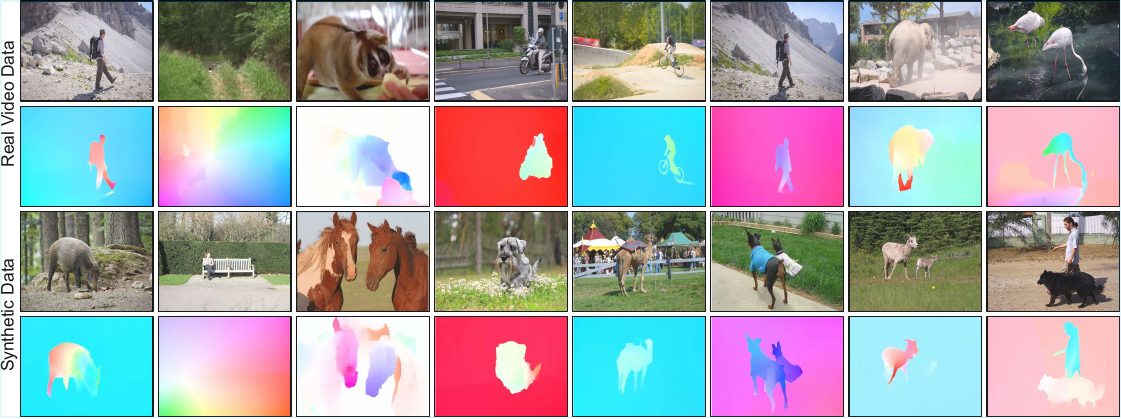}
\vspace{-6mm}
\caption{Aligned qualitative comparison between real video data and our synthetic data.}
\vspace{-1mm}
\label{figure5}
\end{figure*}

\begin{figure}[t]
\centering
\includegraphics[width=1.0\linewidth]{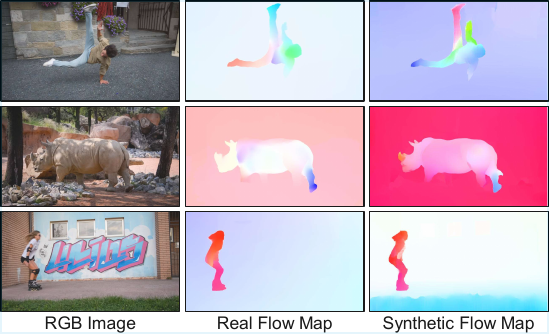}
\vspace{-6mm}
\caption{Direct qualitative comparison of actual and synthetic flow maps, both derived from the same source image.}
\label{figure6}
\end{figure}

\subsection{Analysis}
We conduct experiments to evaluate our approach, focusing on training protocols, data characteristics, architectural choices, and limitations to understand the mechanisms behind performance improvements.

\vspace{1mm}
\noindent\textbf{Training protocol.}
Table~\ref{table3} evaluates different training data configurations. Training exclusively on our DUTS-Video dataset outperforms training on real video data (DAVIS 2016~\cite{DAVIS} and DAVSOD~\cite{SSAV}) across all benchmarks except the DAVIS 2016 validation set. This demonstrates that semantically-aware motion patterns generated through our knowledge transfer approach provide more effective training supervision than limited real video data. By leveraging motion knowledge from pre-trained video diffusion models, our method generates realistic training data that enhances video SOD model development. Combining real video data with DUTS-Video achieves optimal performance across all datasets and metrics, confirming that abundant training data with realistic motion cues significantly enhances video SOD model development.

\vspace{1mm}
\noindent\textbf{Comparison to real data.}
Our approach generates diverse image-flow-mask triplets by creating video sequences from static images and extracting motion patterns from the generated frames. Figure~\ref{figure5} shows that our synthetic data encompass diverse optical flow patterns, from clear object-shaped flows to ambiguous motion representations, providing comprehensive coverage for robust training. Figure~\ref{figure6} further validates our approach through direct comparison with real flows from identical source images, demonstrating that our generated flows capture plausible object movements suitable for video SOD training. This highlights the effectiveness of transferring motion knowledge from video diffusion models to generate semantically-aware optical flows.

\begin{figure}[t]
\centering
\includegraphics[width=1.0\linewidth]{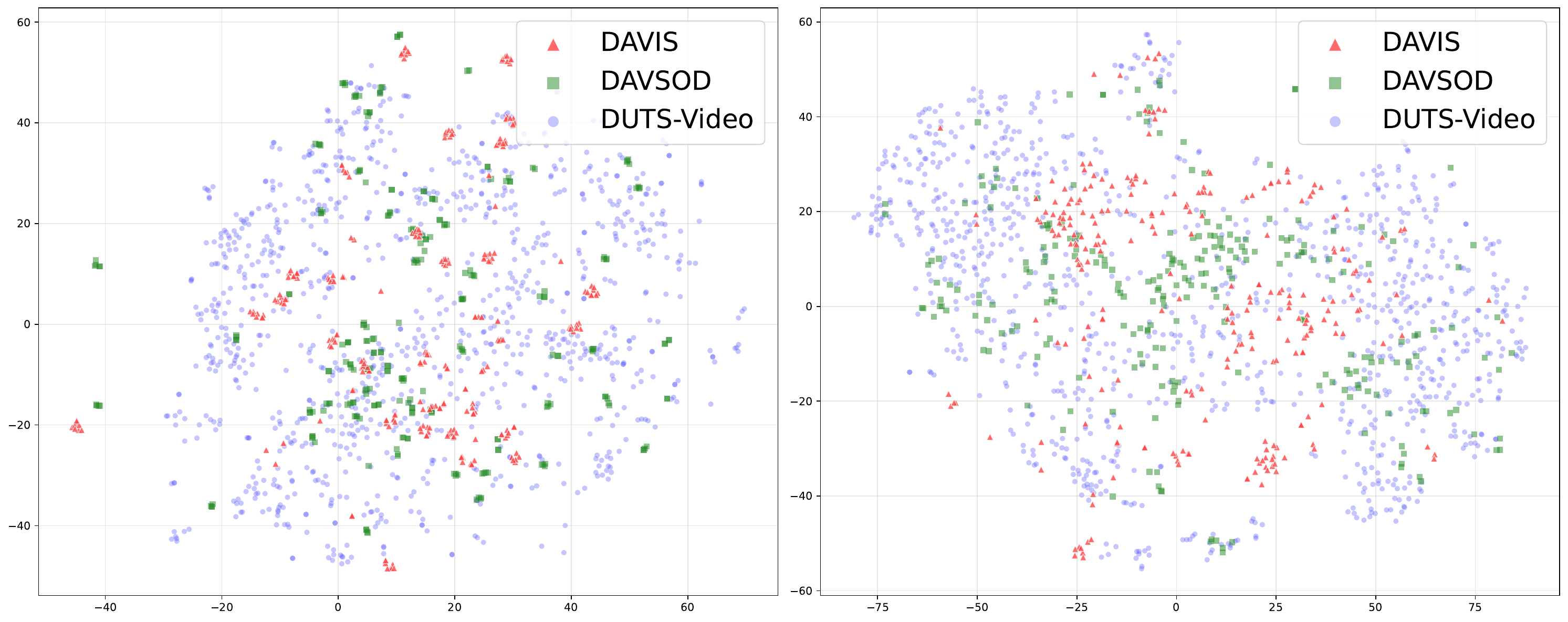}
\vspace{-6mm}
\caption{t-SNE comparison of feature distributions from real and synthetic datasets (left: RGB images, right: optical flow maps).}
\label{figure7}
\end{figure}

\vspace{1mm}
\noindent\textbf{Data distribution.}
Figure~\ref{figure7} visualizes feature embeddings extracted using ResNet-18~\cite{resnet} and projected via t-SNE to compare data distributions across datasets. Real video datasets exhibit distinct, limited distributions, explaining the poor cross-dataset generalization observed in Table~\ref{table3}. In contrast, our synthetic data spans a significantly broader feature space, providing comprehensive appearance and motion diversity that enables robust generalization across various domains and challenging scenarios. This broader feature space reflects the rich semantic motion patterns embedded in our generated training data.

\begin{figure}[t]
\centering
\includegraphics[width=0.8\linewidth]{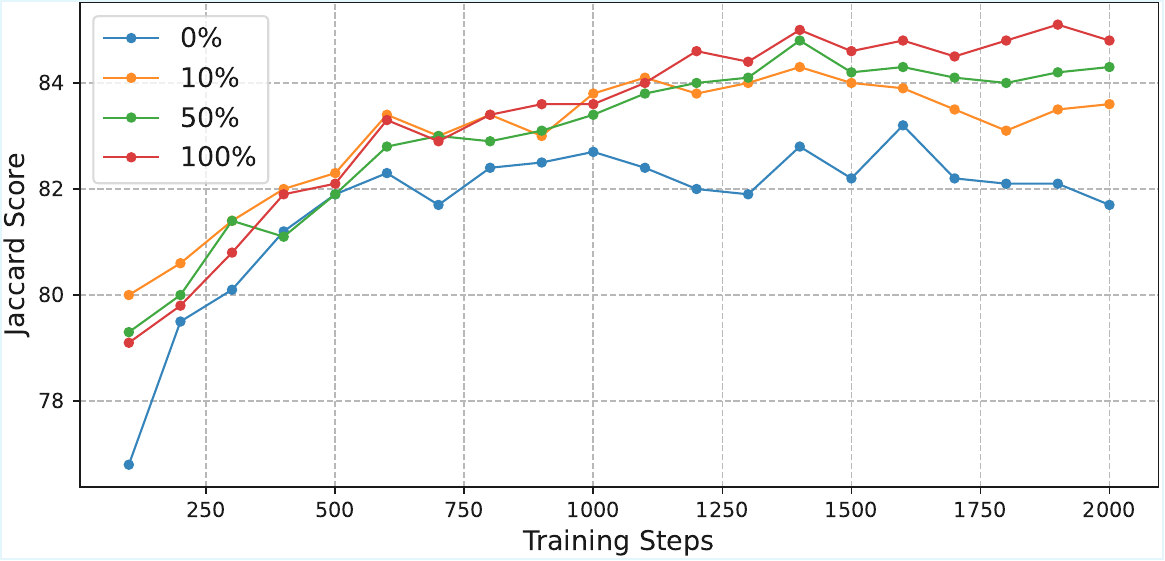}
\vspace{-2mm}
\caption{Jaccard score across training steps for different ratios of synthetic data, evaluated on the DAVIS 2016 validation set.}
\label{figure8}
\end{figure}

\begin{figure*}[t]
\centering
\includegraphics[width=1.0\linewidth]{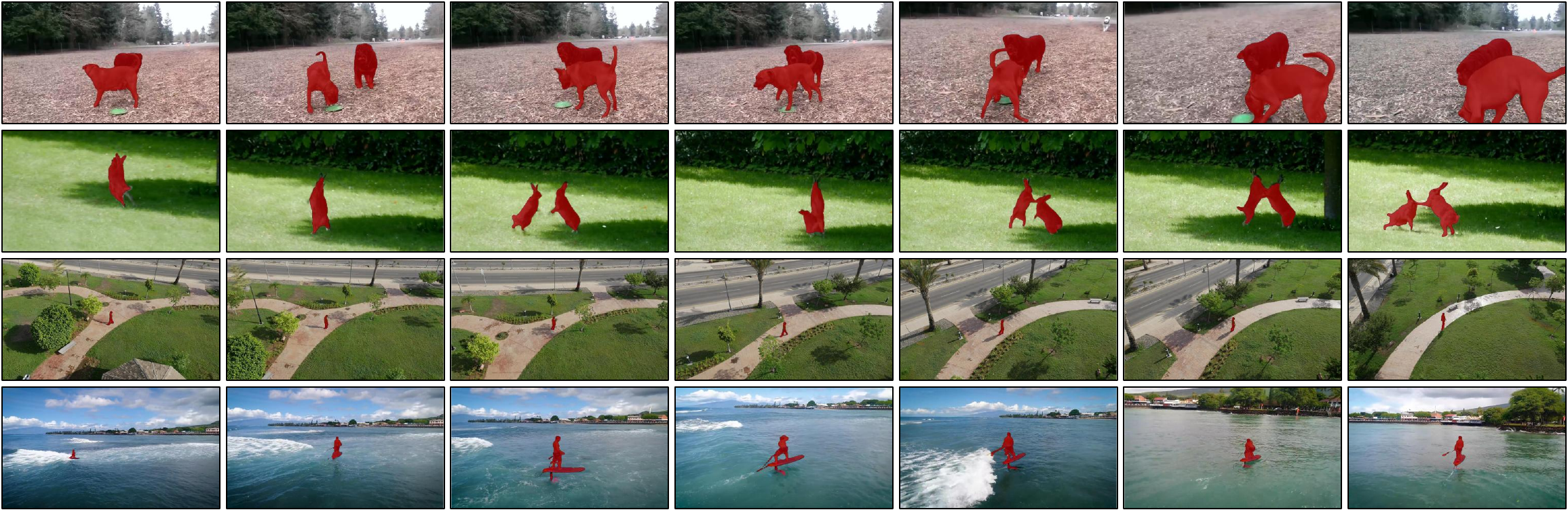}
\vspace{-6mm}
\caption{Qualitative results on the general video cases.}
\label{figure9}
\end{figure*}

\vspace{1mm}
\noindent\textbf{Training statistics.}
We systematically analyze the effect of synthetic training data volume on both convergence speed and segmentation performance. As shown in Figure~\ref{figure8}, models trained with a higher proportion of synthetic data require slightly more training steps to converge, yet consistently achieve superior Jaccard scores compared to those trained with less. These results highlight that, while increased synthetic data may marginally slow convergence, it provides a clear and substantial benefit to overall performance.

\vspace{1mm}
\noindent\textbf{Backbone network.}
Table~\ref{table4} compares different backbone architectures across four benchmarks. Our default MiT-b2~\cite{MiT} backbone achieves optimal performance while maintaining efficient inference speed and parameter count, making it suitable for practical applications. The consistent performance scaling with backbone capacity demonstrates the stability and effectiveness of our training methodology across different architectural choices.

\vspace{1mm}
\noindent\textbf{Input resolution.}
Table~\ref{table5} examines performance-efficiency trade-offs across different input resolutions. Lower resolutions provide faster inference with minimal accuracy degradation, while higher resolutions maximize precision. This flexibility allows adaptation to various computational constraints and application requirements without compromising the core method's effectiveness.

\begin{table}[t]
\centering 
\caption{Ablation study on the backbone network.}
\vspace{-2mm}
\small
\begin{tabular}{P{1.4cm}|P{0.6cm}P{1.1cm}P{0.6cm}P{0.6cm}P{0.6cm}}
\toprule
Backbone &fps &\#Param &$\mathcal{S}\uparrow$ &$\mathcal{F}\uparrow$ &$\mathcal{M}\downarrow$\\
\midrule
MiT-b0 &61.1 &13.8M &89.4 &87.0 &3.5\\
MiT-b1 &45.5 &33.6M &89.8 &87.3 &3.2\\
MiT-b2 &29.5 &55.7M &90.9 &88.6 &2.6\\
\bottomrule
\end{tabular}
\label{table4}
\end{table}

\begin{table}[t]
\centering 
\caption{Ablation study on the input resolution.}
\vspace{-2mm}
\small
\begin{tabular}{P{1.7cm}|P{0.6cm}P{0.6cm}P{0.6cm}P{0.6cm}}
\toprule
Resolution &fps &$\mathcal{S}\uparrow$ &$\mathcal{F}\uparrow$ &$\mathcal{M}\downarrow$\\
\midrule
256$\times$256 &55.0 &90.0 &87.5 &3.0\\
384$\times$384 &44.2 &90.2 &87.8 &2.9\\
512$\times$512 &29.5 &90.9 &88.6 &2.6\\
\bottomrule
\end{tabular}
\label{table5}
\end{table}

\vspace{1mm}
\noindent\textbf{General video cases.}
Figure~\ref{figure9} shows the results on general videos from the MOSE~\cite{mose} validation set and the LVOS~\cite{lvos} test set. We sample scenarios where video saliency is valid. The upper two are from the MOSE validation set, and the lower two are from the LVOS test set. We do not perform additional training on these datasets and directly test the trained model on them. The results demonstrate that TransFlow is effective for handling general video cases, further validating the motion knowledge transfer from video diffusion models.

\begin{figure}[t]
\centering
\includegraphics[width=0.8\linewidth]{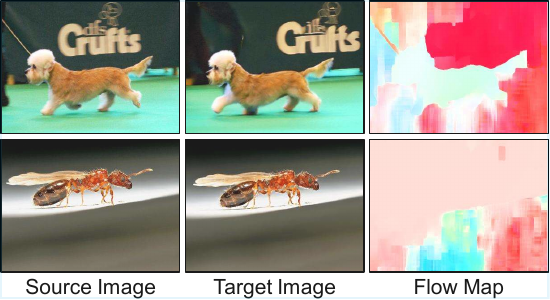}
\vspace{-1mm}
\caption{Visualization of the limitations of our simulated flows.}
\label{figure10}
\end{figure}

\vspace{1mm}
\noindent\textbf{Limitations.}
Our synthetic data generation approach has several limitations. The generated videos occasionally exhibit discontinuous motion dynamics, leading to checkerboard artifacts in the optical flow, as shown in Figure~\ref{figure10}. These artifacts likely arise from patch-level content copying inherent to the diffusion model. Furthermore, since only RGB frames are synthesized, corresponding transformed masks cannot be generated, which limits the diversity of source-target combinations. Addressing these issues presents a promising direction for future work, such as enabling the joint generation of images and masks.

\section{Conclusion}
In this study, we investigate the use of large-scale image data to train flow-guided models. Rather than relying on source image distortion methods that fail to accurately mimic the optical flows of real video data, we demonstrate that novel image generation through image-to-video generation provides an effective solution. For the video SOD task, our approach sets a new state-of-the-art performance across all benchmark datasets, surpassing existing methods.

{
    \small
    \bibliographystyle{ieeenat_fullname}
    \bibliography{TransFlow}
}

\end{document}